\documentclass[10pt,twocolumn,letterpaper]{article}

\usepackage{iccv}
\usepackage{times}
\usepackage{epsfig}
\usepackage{graphicx}
\usepackage{amsmath}
\usepackage{amssymb}

\usepackage[pagebackref=true,breaklinks=true,letterpaper=true,colorlinks,bookmarks=false]{hyperref}

\iccvfinalcopy 

\def\iccvPaperID{7051} 

\ificcvfinal\fi

\usepackage{color,xcolor}
\newcommand{\shuran}[1]{#1} 
\newcommand{\cheng}[1]{#1} 
\usepackage{enumitem}
\usepackage{tabularx}
\usepackage{booktabs}
\usepackage{soul}

\makeatletter
\DeclareRobustCommand\onedot{\futurelet\@let@token\@onedot}
\def\@onedot{\ifx\@let@token.\else.\null\fi\xspace}

\makeatother

\definecolor{MyDarkBlue}{rgb}{0,0.08,1}
\definecolor{MyDarkGreen}{rgb}{0.02,0.6,0.02}
\definecolor{MyDarkRed}{rgb}{0.8,0.02,0.02}
\definecolor{MyDarkOrange}{rgb}{0.40,0.2,0.02}
\definecolor{MyPurple}{RGB}{111,0,255}
\definecolor{MyRed}{rgb}{1.0,0.0,0.0}
\definecolor{MyGold}{rgb}{0.75,0.6,0.12}
\definecolor{MyDarkgray}{rgb}{0.66, 0.66, 0.66}

\newcommand{\mypara}{\vspace{-3mm}\paragraph}
\newcommand{\OURS}{GarmentNets\xspace}

\begin{document}

\title{\OURS: Category-Level Pose Estimation for Garments via \\ Canonical Space Shape Completion\vspace{-3mm}}

\author{Cheng Chi \quad \quad Shuran Song \\
Columbia University\\
\href{https://garmentnets.cs.columbia.edu}{https://garmentnets.cs.columbia.edu} \vspace{-3mm}
}

\maketitle
\ificcvfinal\thispagestyle{empty}\fi

\begin{abstract}
This paper tackles the task of category-level pose estimation for garments. 
With a near infinite degree of freedom, a garment's full configuration (i.e., poses) is often described by the per-vertex 3D locations of its entire 3D surface. 
However, garments are also commonly subject to extreme cases of self-occlusion, especially when folded or crumpled, making it challenging to perceive their full 3D surface.  
To address these challenges, we propose GarmentNets, where the key idea is to formulate the deformable object pose estimation problem as a shape completion task in the canonical space. This canonical space is defined across garments instances within a category, therefore, specifies the shared category-level pose. By mapping the observed partial surface to the canonical space and completing it in this space, the output representation describes the garment's full configuration using a complete 3D mesh with the per-vertex canonical coordinate label. 
To properly handle the thin 3D structure presented on garments, we proposed a novel 3D shape representation using the generalized winding number field. 
Experiments demonstrate that GarmentNets is able to generalize to unseen garment instances and achieve significantly better performance compared to alternative approaches. Code and data can be found in \href{https://garmentnets.cs.columbia.edu}{https://garmentnets.cs.columbia.edu}

\end{abstract}

\section{Introduction}
\vspace{-2mm}
Garments are one of the most common objects in our life, yet they pose a set of unique properties that make them incredibly difficult for machines to perceive and interact: 
\begin{itemize}[leftmargin=*]
\vspace{-2mm}
    \item {Infinite degree of freedom (DoF)}: in contrast to rigid objects whose pose can be fully specified as a low-dimensional vector, a piece of garment has near infinite DoF, i.e., to fully specify its configuration (i.e., pose), we need to describe positions of all 3D points on the garment surface. This issue is compounded when we consider category-level generalization, where there are infinite poses can be considered as the ``canonical'' for different garments instances.

\vspace{-2mm}
    \item {Severe self-occlusion}: garments are often subject to extreme cases of self-occlusion, especially when folded or crumpled. This property makes it particular challenging and sometimes ambiguous to perceive their full configuration from partial visual observation.  
    
\vspace{-2mm}    
    \item {Thin structure}: garments often consist of thin 3D geometric structures that are not water-tight. This unique geometric property makes them ill-suited for typical 3D shape representations designed for solid rigid objects (e.g., occupancy grid or signed distance functions). 
\vspace{-2mm} 
\end{itemize}

\begin{figure}[t]
\centering
\includegraphics[width=\linewidth]{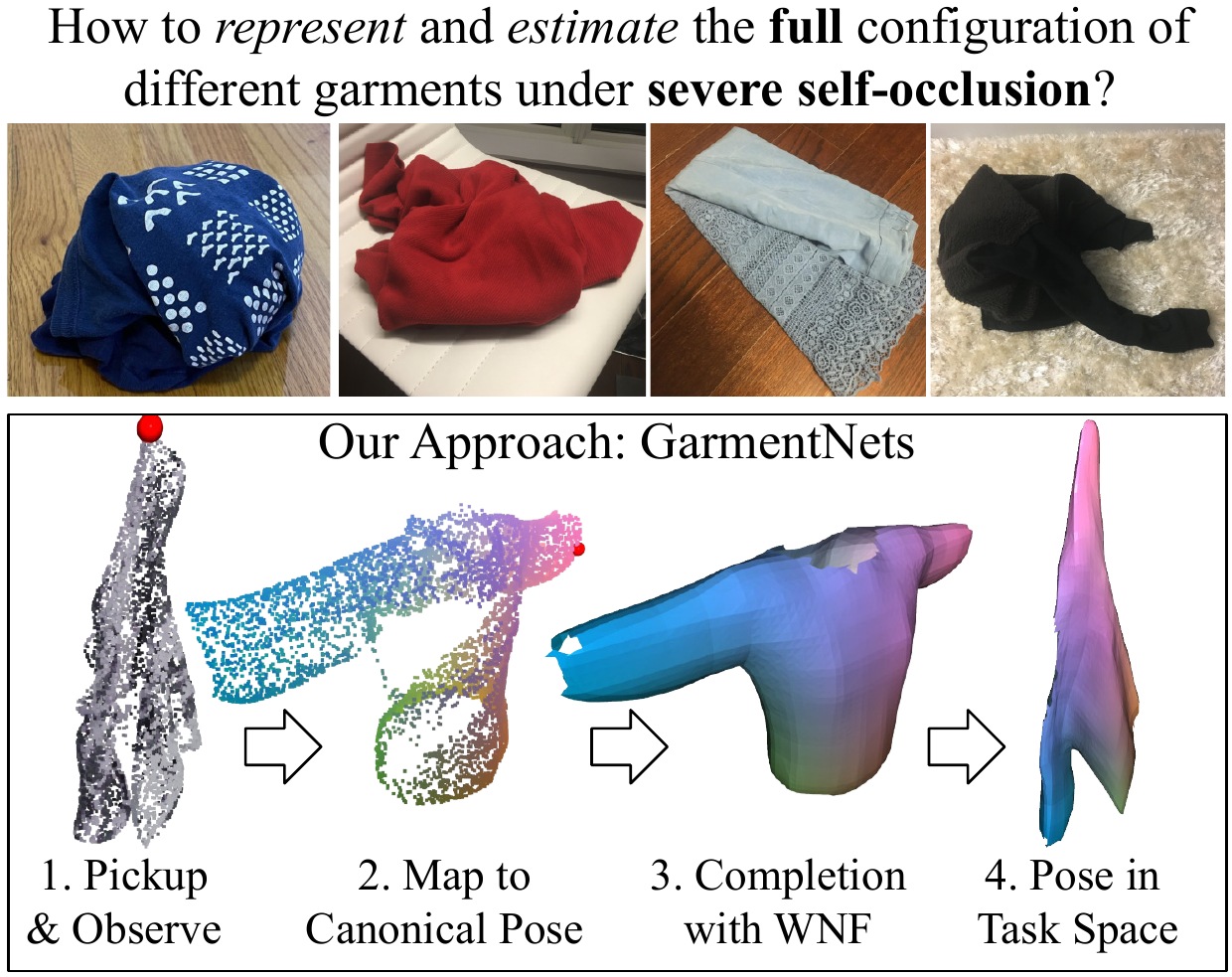}
\caption{
\textbf{Category-level Pose Estimation for Garments.} The key idea of \OURS is to formulate the garment pose estimation problem as a shape completion task in the canonical space. This canonical space is defined across garment instances within a category, therefore, specifies the shared category-level pose. The output describes the garment's full configuration using a complete 3D mesh with per-vertex canonical coordinate label.}
\vspace{-5mm}
\label{fig:teaser_figure}
\end{figure}

Due to these challenges, prior works on garments or cloth perception often build on simplifying assumptions such as full visibility \cite{dynamic_fusion}, known instance-level mesh \cite{peter_alan_baseline, cpd_physics, cpcpd}, known physics or full state information in the initial observation \cite{cpd_physics}, where the problem is reduced to an instance-level tracking task.  As a result, these algorithms cannot generalize to new garment instances that are not observed during training. 

To address these challenges, we propose \textbf{\OURS}, an end-to-end neural network that estimates the full configuration of a garment from a partial observation.
The algorithm highlights following key ideas:  

To handle the infinite DoF and enable category-level generalization, we define a \ul{normalized coordinate space} for each garment category using a canonical human pose.  This representation allows the algorithm to learn semantically meaningful correspondences between garment instances with different styles, shapes, or configurations.

To handle self-occlusions, the algorithm explicitly performs \ul{shape completion} under its canonical pose, which allows the algorithm to fully specify the garment configuration even when the observed surface is incomplete.  

To handle thin structures, we propose a novel 3D shape representation using \ul{winding number field} (WNF) \cite{jacobson2013robust}. This representation allows the algorithm to accurately represent thin cloth structures with strong gradient on the surface but continuous and smooth elsewhere, providing a more meaningful signal for the network to learn better geometric features.

We study the garments perception task in the context of \textit{robot manipulation}, which is more challenging than the \textit{on-body} garments perception (i.e., the garment being worn by people) since the number of possible configurations is much larger and the potential self-occlusion is more severe. 
However, this setup also allows us to leverage simple robot interactions to reduce the possible configuration space.
For example, we allow the robot to first lifts a crumpled garment with a random pick point and allows the gravity force to naturally pulls the cloth into a stable pose. The system then takes four RGB-D images of the cloth by rotating the gripper. This task formulation potentially allows our perception algorithm to be used in a realistic robot manipulation task. 

To the best of our knowledge, we are the first to enable category-level full configuration estimation of garments from partial observations. 
Our experiments demonstrate that the trained model is able to generalize to \textit{novel} garment instances as well as \textit{real world} images.

\section{Related Work}
\vspace{-2mm}
\paragraph{Pose estimation.}
Pose estimation for rigid objects is extensively studied in both computer vision and robotics community \cite{Wang_2019,Zakharov_2019,wang2019densefusion,Tekin_2018,Xiang_2018}, where the task is to predict a 6 DoF vector that describes the object rotation and translation given the visual observation(s) of a known object instance. 
He et al. \cite{nocs} extended this task definition to category-level by defining a normalized canonical space (NOCS) for different object instances within a category and allowing the algorithm to infer \cheng{scale}.
Recently, Li et al. \cite{Li_2020_CVPR} further extended this approach to handle category-level articulated objects by defining additional canonical joint configurations.
However, different from rigid or articulated objects, a garment has near-infinite DoF and violates the pieces-wise rigid body assumption, \shuran{which makes them inapplicable for these approaches.}
\vspace{-2mm}

\mypara{Shape completion.}
Many shape completion algorithms are proposed for rigid objects using different 3D representations, such as occupancy grid \cite{shape_completion_encoder},  distance function \cite{dynamic_fusion, r1_chibane2020ndf}, point cloud \cite{pcn} and implicit function \cite{conv_implicit}. In particular, implicit function  has been a popular solution to increase prediction resolution while maintaining low memory consumption. \cheng{However, garments' high intra-class shape variation and thin geometry structure motivates a new class of shape representations. We proposed a novel winding number field representation to address these concerns.}
\vspace{-2mm}

\mypara{Instance-level cloth perception.}
Vision algorithms for deformable objects have been mostly focused on instance-level tasks.
In 3D reconstruction, a series of work considers deformable object from a pure geometric perspective without leveraging any semantic priors \cite{whelan2016elasticfusion,innmann2016volumedeform}. For example, DynamicFusion \cite{dynamic_fusion} warps and accumulates different observations into a canonical pose (i.e., first frame) as a 3D volume. \shuran{However, it is unable to reconstruct region that occluded throughout the sequence. In constrast, our method can infer occluded parts of cloth with severe deformations by leveraging category-level semantic priors.} 
In pose estimation, most of the prior works simplify the task by assuming known initial state \cite{cpcpd}, instance-level 3D mesh \cite{deepgarment}, or additional visual markers \cite{marked_cloth}. \cheng{Our approach does not rely on the above assumptions and instead use simple robot interactions (i.e., random pickup) to reduce the possible configuration space.}

\vspace{-2mm}
\mypara{On-body cloth perception.}
\cheng{Learning methods have found great success on clothed human reconstruction by leveraging human body shape prior \cite{r1_jiang2020bcnet,r1_saito2020cvpr,r3_patel20tailornet,clothcap,deep_fasion3d}. However, explicitly representing garments on top of an articulated body shape model such as \cite{smpl} makes these model unable to express the highly crumpled garment shapes common in robotic manipulation tasks. Bhatnagar et.al. \cite{r1_bhatnagar2020loopreg} performs registration only, without shape completion. \cite{r1_bhatnagar2020ipnet} performs shape completion directly in task space. Combined with pose estimation, our method is able to perform shape completion in the canonical space, which is more robust to large deformation and occlusions.}

\begin{figure*}[t]
\begin{center}\vspace{-5mm}
\includegraphics[width=\linewidth]{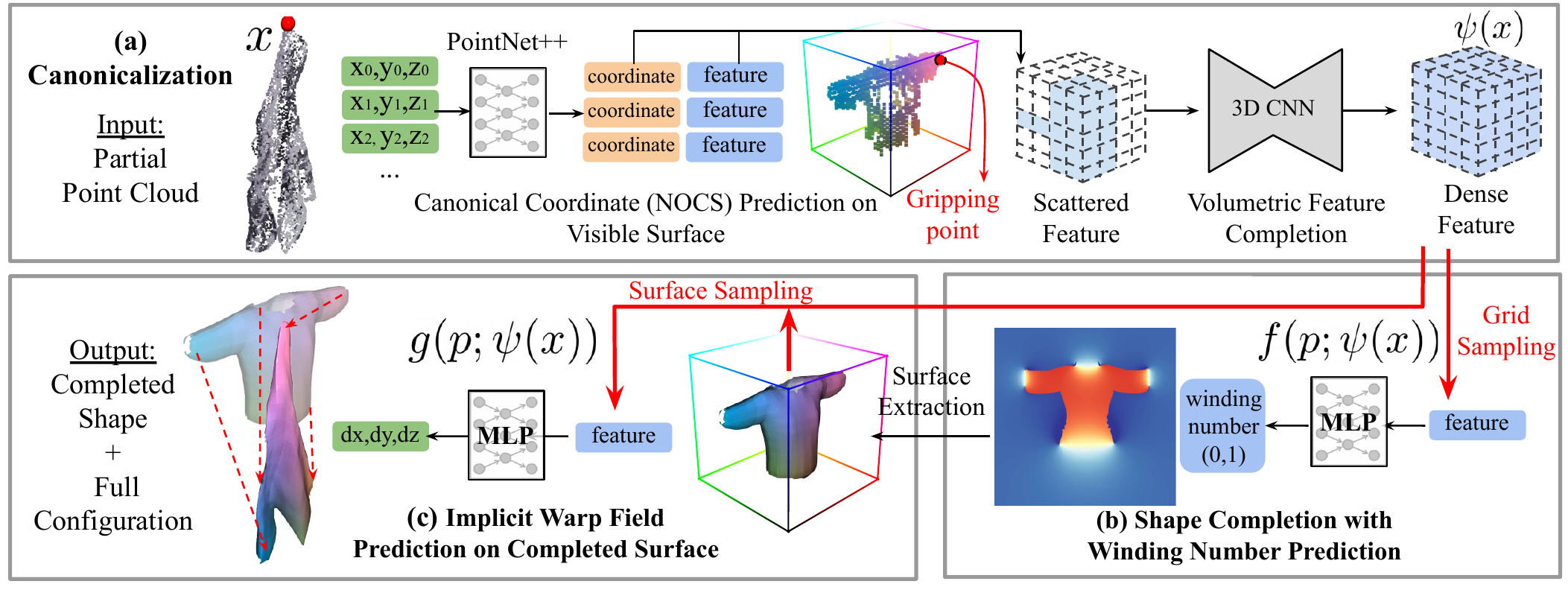}
\end{center} \vspace{-5mm}
\caption
{
\textbf{Network Overview.} Given a colored point cloud of a grasped garment in the task space, the \textbf{(a)} NOCS network predicts canonical coordinates for the observed points. The predicted coordinates are used to scatter the point-wise features into a 3D volume by providing the position index in the target volume. The scattered sparse feature volume is passed to a 3D CNN to produce a dense feature volume. Then the \textbf{(b)} shape completion network infers the garment's full 3D geometry by predicting a winding number field for sampled positions $p$. Finally, the \textbf{(c)} warp field network  predicts an implicit warp field that maps the completed surface (in canonical pose) back to original task space. 
The output mesh then encodes the garment’s full configuration using per-vertex canonical coordinate label. }
\label{fig:pipeline_figure} \vspace{-4mm}
\end{figure*}
\section{Approach}
With a near-infinite degree of freedom, a garment's full configuration (i.e., pose) is often described by its pre-vertex 3D locations, which is ill-defined without knowing the full cloth geometry. On the other hand, only reconstructing the 3D geometry of the garment does not provide semantically meaningful correspondence between different garment instances (e.g., where is the sleeve?), which are useful for many applications (e.g., folding).

In this paper, we propose to formulate the deformable object pose estimation problem as a shape completion task in the canonical space.  Given a partial point cloud of an unseen garment, \OURS first maps the observed points to a category-level canonical space, completes the garment's 3D geometry in this space, and finally warps it back to the observation space. The output configuration is described by a completed 3D mesh where each vertex is labeled with its corresponding coordinate in the canonical space.
This canonical space defines a shared category-level pose by specifying a semantically meaningful correspondence across different garments instances within a category. 
The following sections provide details for key algorithm components and design decisions.

\subsection{Pick First then Recognize\label{sec:pick}}
Perceiving the full configuration of a garment in its arbitrary crumpled state (like examples in Fig. \ref{fig:teaser_figure}) is extremely challenging and oftentimes impossible. 
Inspired by Li et al. \cite{peter_alan_baseline}, we make use of simple robot interactions to reduce the possible configuration space of the garment and increase its visibility. The robot first lifts the crumpled garment with a random pick point and shakes it to allow the gravity force to pull down the cloth into a stable pose naturally. 
After the robot's gripper lifts the garment, one or multiple RGB-D images were taken and converted into an RGB point cloud $x$ in the gripper frame (i.e., the gripping point is the origin). After that, \OURS is trained to estimate these garments' poses under different gripping positions from their point cloud observations. 

This ``pick up first, ask questions later'' strategy is widely used in robot perception \cite{zeng2018robotic, peter_alan_baseline,xiao2005uncalibrated}, where the robot can reliably grasp the object without recognizing the pose of the object. This assumption is valid for most garments since almost any surface point on the garments is graspable by the robot. Meanwhile, picking up the garment helps in isolating it from clutter, increase its surface visibility which is all beneficial for the downstream perception algorithm. 

\subsection{Normalized Canonical Space for Garments \label{sec:nos}}
A dense, semantically meaningful label is helpful for downstream tasks such as folding or pick and place. Manually specifying such label can be expensive and time-consuming \cite{dense_pose}. Here, we extended He et al.\cite{nocs}'s framework by using the point locations in a categorically normalized space as a per-point label, which naturally provides intra-category correspondence.

\begin{figure}[t]
       \centering
       \includegraphics[width=\linewidth]{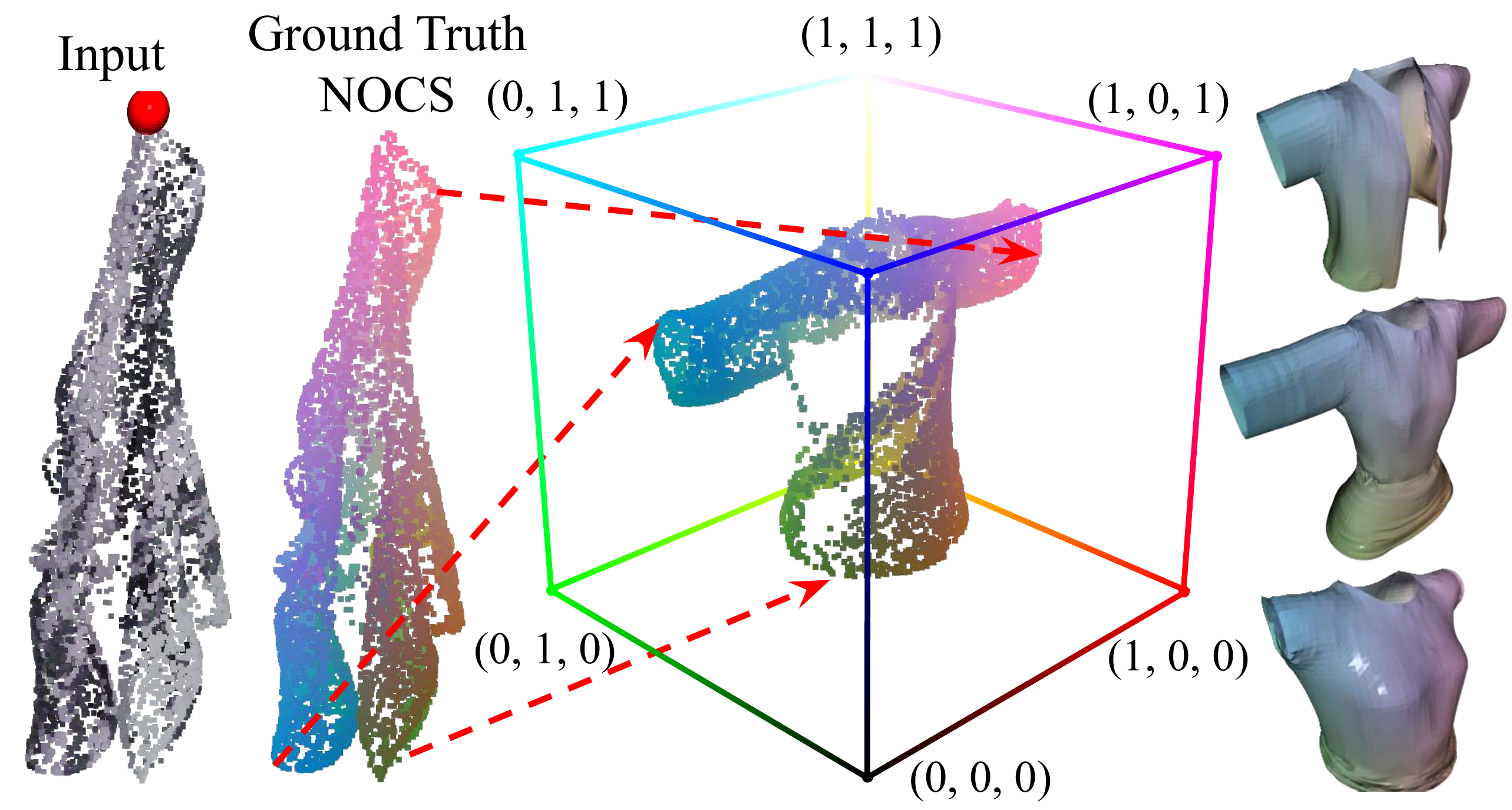}
       \caption{\textbf{Canonical Coordinate (NOCS) for Garments.} The canonical coordinate is defined as the garment in canonical configuration scaled into a unit cube using a per category scaling factor.}
       \label{fig:nocs}
       \vspace{-4mm}
 \end{figure}
\mypara{Definition} 
A garment's canonical space (i.e., NOCS) is defined by simulating the garment worn by a human in T-pose (provided by CLOTH3D \cite{cloth_3d}). With the human pose specified using SMPL \cite{smpl}, the axis-aligned bounding box for all instances in a category is computed. All instances within each category are then transformed with the same scale and translation such that the largest dimension of the bounding box fits a unit cube (Fig. \ref{fig:nocs}). 
Note that our algorithm does not depend on the specific definition of the canonical pose. For other objects, any canonical pose that provides high surface visibility can be chosen.

\mypara{Canonical Coordinate Prediction.}
Given a colored point cloud observation of the garment, we use PointNet++ \cite{pointnet2} based network to predict a per-point canonical coordinate.  
We formulate this prediction as a classification task by dividing each axis into 64 bins, where the network predicts each axis independently.
We found this classification formulation is much more effective than regression since it allows the network to model the bimodal distribution of coordinate prediction caused by symmetry. In contrast, L2 regression loss encourages the network to predict the mean between the two hypotheses (Fig. \ref{fig:regression_classification}).
While the bin size limits the prediction accuracy, this step's primary goal is to scatter the per-point feature vector to roughly the correct location in the feature volume, which is sufficient with the current resolution. 
Although the network does make mistakes in this stage due to ambiguities in symmetry (e.g., predicting left sleeve as right), we observe that the latter stages of the network (shape completion and warp field prediction) are able to correct some of these errors through learning.
The network is trained with a ground truth NOCS label using CrossEntropy Loss, and its weights are fixed during shape completion and warp field module training. During training, the input point cloud is randomly downsampled into 6000 points and augmented with a random Z rotation around the gripping point.

\mypara{Feature Scattering using Canonical Coordinates.}
After obtaining the coordinate prediction for each observed 3D point, we ``scatter'' the per-point feature vector into a $32^3$ feature volume.
The ``feature'' being scattered is a concatenation of the original 3d coordinate of the point, the predicted canonical coordinate, the confidence of for NOCS prediction on each dimension, and the second to last layer 128-dimensional PoinetNet++ feature, in total 137-dimension. 
This concatenated feature is passed through an MLP (multi-layer perceptron) before aggregation.
The ``scatter'' operation is performed by copying the feature vector to the target volume location using the predicted NOCS coordinate. All features mapped to the same volume index will be aggregated using a channel-wise maximum. 
The feature vectors with no corresponding input points are initialized with zeros.
 This aggregated sparse feature volume is further transformed with a 3D UNet \cite{unet_3d} to generate a dense feature volume $\psi(x)$.
The weights of shared MLP and 3D Unet are trained with shape completion (Sec. \ref{sec:completion}) and warp field prediction (Sec. \ref{sec:warp}) modules jointly.

\begin{figure}[t]
       \centering
       \includegraphics[width=\linewidth]{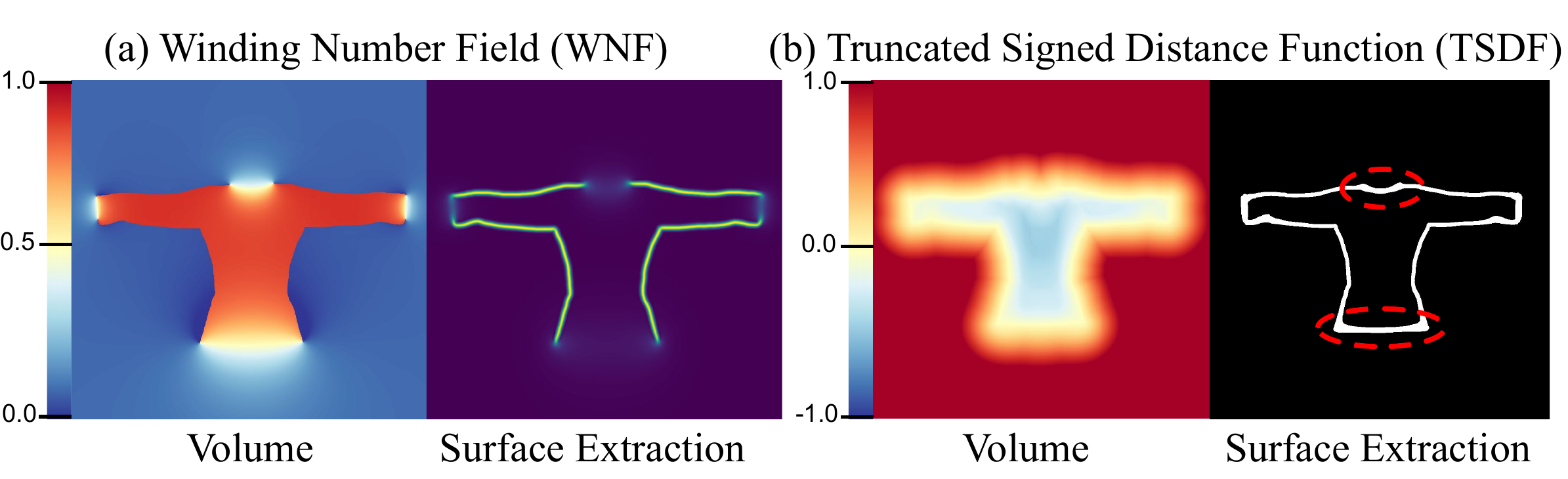}
       \caption{WNF and TSDF computed using the canonical mesh. Note that TSDF's zero-crossing surface also includes waist, neck and sleeve, which is undesirable. In contrast, WNF can differentiate surface and openings using the magnitude of gradient.}
       \label{fig:gt_wnf}
       \vspace{-4mm}
 \end{figure} 

\subsection{Shape Completion with Winding Number \label{sec:completion} } 
At this stage,  canonical coordinates are predicted for all visible points with limited resolution. To estimate the pose of the occluded surface, we perform a volumetric shape completion in the canonical space. 
Shape completion for rigid objects is a well-established task with many prior works.
In this work, we borrow the successful techniques and ideas from the related fields. 
However, the non-watertight thin structure of garments raises a unique challenge for this task.  

Typical 3D shape representations, such as occupancy grids, will be very challenging to represent the thin surface accurately. 
The accuracy is limited by its voxel resolution; however, high resolution volume will result in extremely sparse data distribution, and high memory consumption.
Using Truncated Signed Distance Function (TSDF), we are able to represent accurate thin surface structure by giving different signs for voxels inside or outside the garment (Fig. \ref{fig:gt_wnf}), where the zero-crossing surface precisely describes the location of the surface. However, this TSDF representation also creates additional artificial surfaces around garment openings (e.g., neck, waist) due to the change of signs.

\mypara{Winding Number Field for Shape Representation.} To address this issue, we adopt the Generalized Winding Number proposed by Jacobson et al. \cite{wnf} as the shape representation. For a point $p \in \mathbf{R}^3$ and a surface $S$, the generalized winding number is defined by integrating the solid angle over the surface.
$ w(p)=\frac{1}{4\pi} \iint \limits_{S}sin(\phi) d\theta d\phi$.

Intuitively, if the surface $S$ is watertight and has no self-intersection, the winding number equals $1$ if $p$ is inside $S$, and $0$ if outside. However, if $S$ is not watertight, Jacobson et al. has proved that the winding number will be a harmonic function with the boundary condition that the inner side of the surface equals $1$ and the outer side equals $0$. When crossing the surface directly, the winding number jumps from $1$ to $0$. When crossing a surface opening, the winding number smoothly transitions from $1$ to $0$ in a way that minimizes the Dirichlet energy. This property of the winding number field allows us to represent whether a point on the watertight implicit surface $w(p)=0.5$ is actually on the non-watertight surface $S$ using the spatial gradient, as shown in Fig.  \ref{fig:gt_wnf}. 

In practice, computing the generalized winding number field for a triangular mesh will require summing the solid angle over all triangles for each query point, which is prohibitively expensive. We use the algorithm proposed by Barillet al. \cite{fast_wnf} to accelerate the computation.  
This representation is friendly for deep learning since it provides a strong gradient on the surface but continuous and smooth elsewhere.
To the best of our knowledge, this is the first time that the winding number field is used in deep learning.

\mypara{Shape Completion Network.} To enable high resolution prediction with reasonable memory consumption,  we use a network structure that combines the 3D CNN and implicit neural representation, inspired by \cite{conv_implicit}. 
Given the dense feature $\psi(x)$ is produced by the 3D CNN network described in Sec. \ref{sec:nos},
the shape completion network predicts the winding number field as a neural implicit function $w(q)=f(q;\psi(x))$, where  $q$ is a query point in 3D space. 
For each query point $q$, we first trilinear interpolate the $32^2$ dense feature volume to get the feature at this point $\psi(p;x)$. 
Then, this feature is concatenated with the query point and transformed by an MLP, which outputs a single scalar as winding number prediction $w(q)$. 
Then a watertight triangular mesh is extracted using marching cubes from the predicted winding number field. The magnitude of the spatial gradient is evaluated for each vertex of the mesh. A constant threshold is used to determine whether the vertex belongs to the surface or an opening.  

\mypara{Training and Inference.} During training, $6000$  query points are uniformly sampled for each instance. The network is trained with L2 loss. Note that the gradient from the MLP will be propagated to the 3D UNet. During prediction, $f(q;\psi(x))$ will be evaluated for all sample points to generate the final winding number field volume. For simplicity, we directly predicted a dense $128^3$ winding number field volume by slicing it into 8 $64^3$ volumes. 

\subsection{Canonical to Task Space Mapping \label{sec:warp}}
Finally, we would like to map the predicted mesh from the canonical space back to the task space (i.e., the coordinate frame of the original input point cloud, where the grasp point is origin). 
This output informs the robot about the full configuration of the garment in the observation space, including the occluded parts. 

\mypara{Phsyics Simulation.} One possible approach is to physically simulate the pose of predicted canonical mesh with the predicted grasp point, assuming that the physical parameters of the garment are known. 
Since we now know the grasp location in the input point cloud (i.e., origin), we can infer the grasp point on the predicted mesh by using the canonical coordinate prediction of the observed point that is closest to the origin. 
By simulating the physical process of 3D mesh being gripped by the grasp point, we can map the prediction back to task space.  
Since the simulated result will be ambiguous up to a rotation around the gravity vector, we can compute the optimal rotation alignment by minimizing its Chamfer distance with respect to the input point cloud. 
However, as shown in our experiments, this approach does not yield the best quantitative result due to its sensitivity to incorrect gripping point prediction or mesh reconstruction. The assumption about known physical parameters also limits its applicability. 
Therefore we also propose to directly infer the per-vertex warp field using a neural implicit  function.

\mypara{Implicit Warp Field Prediction Network.} 
In this approach, we predict the warp field as another implicit neural function $g(p;\psi(x)) \in \mathbf{R}^3$ that takes in a sample position $p$ and infers the warp field for that position (i.e., its task space location). 
In practice, while $g(p;\psi(x))$ is defined for all points in $p \in \mathbf{R}^3$, we can only obtain its ground truth value on the garment surface. If the canonical space mesh is predicted with error, we rely on the neural network's generalizability to obtain the warp field prediction. 

Note that in the feature scattering step, the observed point location is included as part of the feature that is delivered into the sparse feature volume.
Leveraging this information, the network gains additional robustness against mirrored predictions. For example, if the point belongs to the left sleeve is predicted on the right side of canonical space, it's task space coordinate and features will also be delivered to the right side of of the feature volume, which can make $g(p;\psi(x))$ to predict the left sleeve's task pace coordinate for that point $p$, resulting in correct location.

\begin{figure*}[t]
    \centering \vspace{-5mm}
    \includegraphics[width=0.96\linewidth]{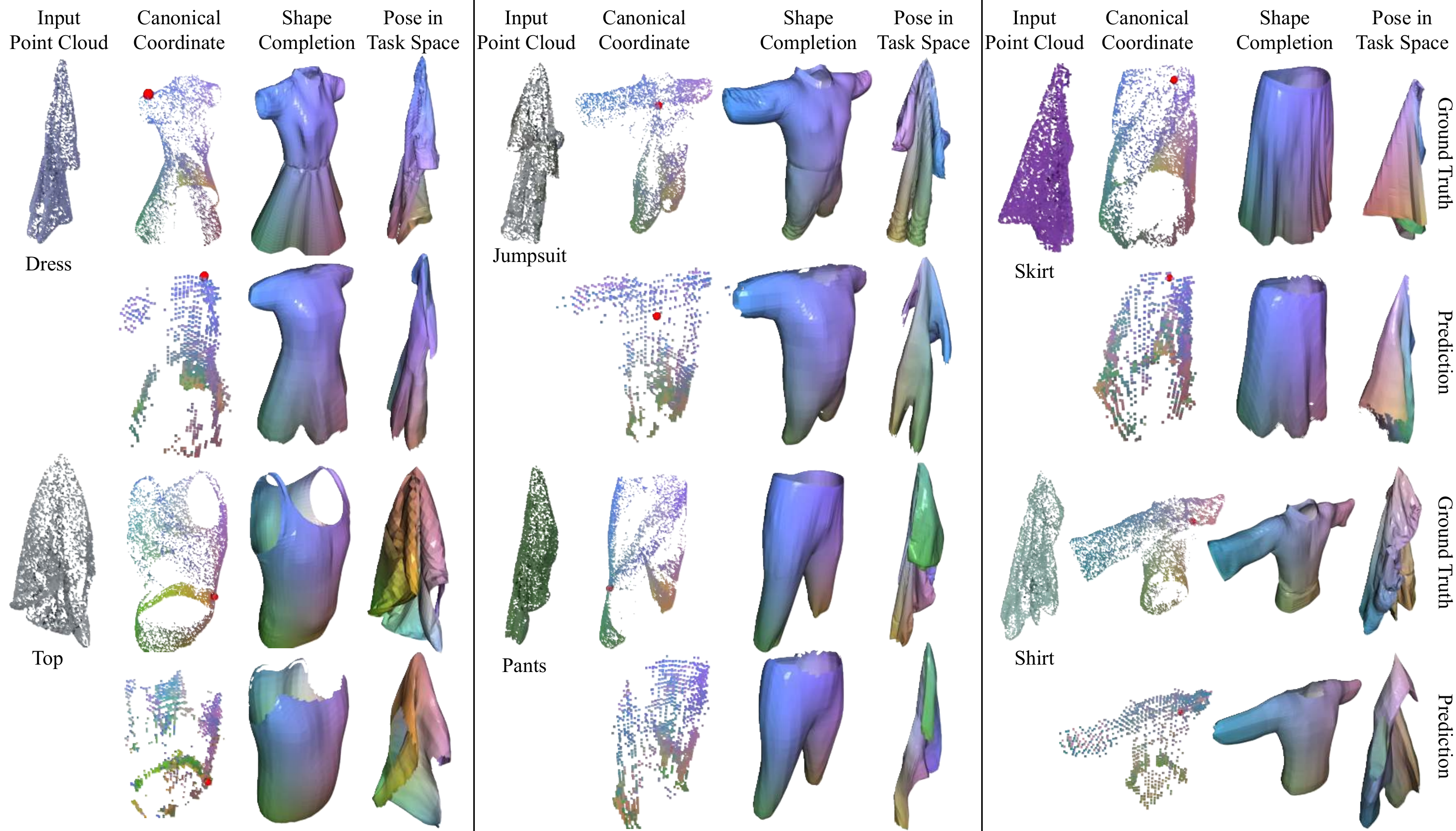}
    \caption{\textbf{Qualitative Results on Unseen Garment Instances (Simulation).} From left to right shows the input and output of each stage. The ground truth and predicted grasp points are shown as red spheres. Note that despite that the predicted gripping point on the Dress and Trousers example are on the wrong side, the final pose were still predicted correctly, thanks to the warp field prediction. }
    \label{fig:result}
    \vspace{-5mm}
\end{figure*}

\mypara{Training and Inference.} During training, 6000 points are uniformly sampled on each ground truth mesh surface. The warp field network is trained with L2 loss. The winding number field and warp field modules are trained simultaneously using a shared feature volume  $\psi(x)$. The loss of the two modules are added with equal weight. During prediction, each vertex in the predicted canonical mesh surface is used as a query point to predict task space coordinate.

\section{Evaluation}
 
\noindent \textbf{Data Generation.} We use canonical pose meshes from the CLOTH3D dataset \cite{cloth_3d} to generate our data. The dataset has six garment categories. We simulated each garment instance 21 times using a randomly sampled gripping point. We used Blender to simulate the physics and render the ground truth images (i.e., RGB-D images, UV maps and object masks).   The training, validation, and testing set are disjoint at the garment instance level. 

\mypara{Metric:} We use the following two metrics for evaluation:
\begin{itemize}[leftmargin=*]
\vspace{-1mm}
    \item \textbf{Symmetric Chamfer Distance ($D_c$).} This metric measures accuracy and completeness for \textit{surface reconstruction}. The accuracy metric is defined as the mean L2 distance of points on the output mesh to their nearest neighbors on the ground truth mesh. The completeness metric is defined similarly but in the opposite direction. We estimate both distances efficiently by randomly sampling 10k points from both meshes and using a KD-tree to estimate the corresponding distances. 
    $D_c$ is measured in task space for Tab. \ref{tab:baselines} and in canonical space for Tab. \ref{tab:supp_reconstruction}. 

\vspace{-2mm}
    \item \textbf{Correspondence Distance ($D_n$).} Similar to $D_c$, we compute point-wise L2 distance between the predicted surface and the ground turth surface. However, the correspondences are established using the closest point between predicted and ground truth NOCS labels instead of the closest point in 3D. 
This metric measures the \textit{pose estimation} accuracy. 
\vspace{-2mm}
\end{itemize}

\section{Experimental Results}
\begin{figure}[t]
    \includegraphics[width=\linewidth]{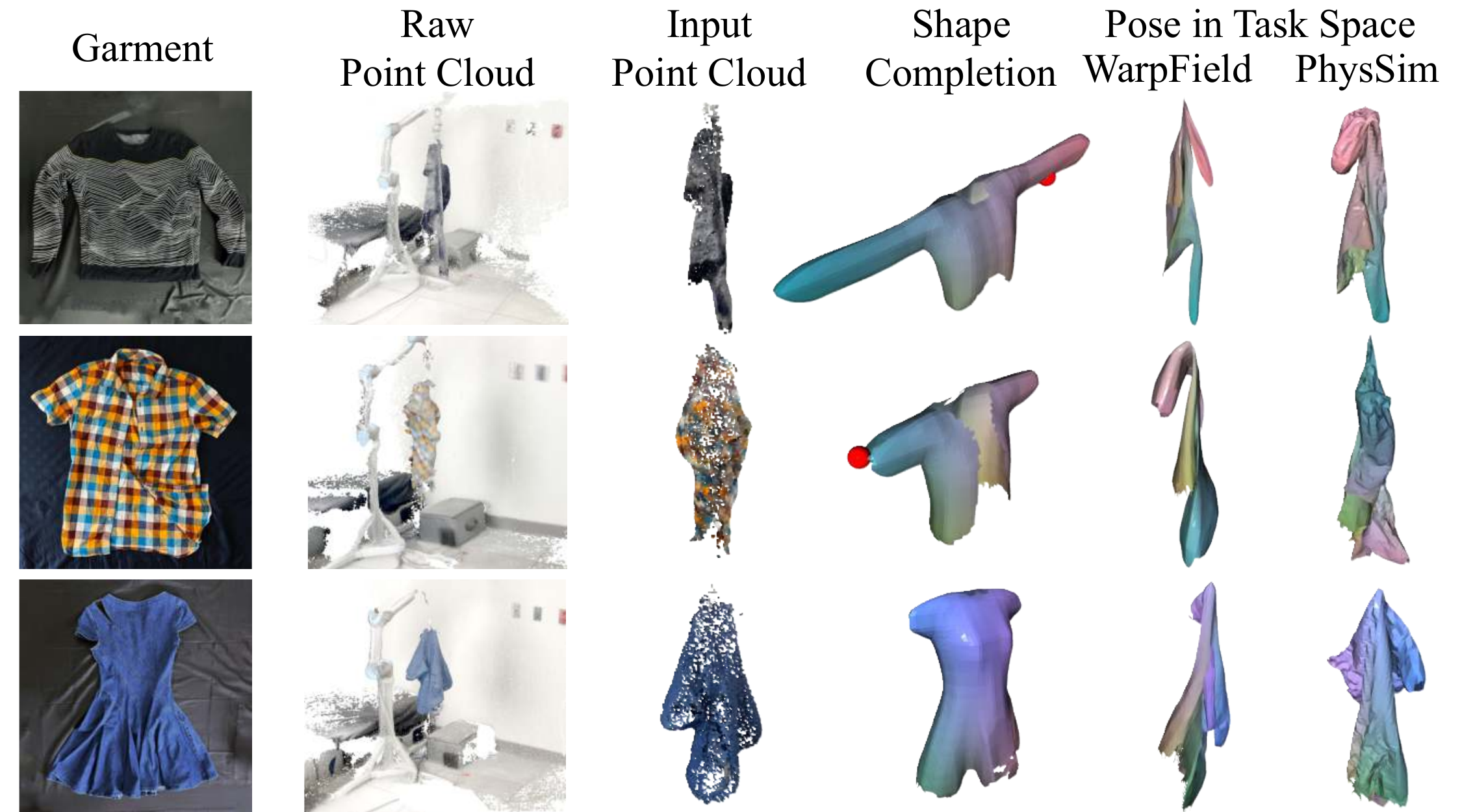}
    \caption{\textbf{Qualitative Results on Unseen Garment Instances (Real World Data).} We validate our algorithm on realworld garments where the garments is lifted by the robot arm and the point cloud captured with iPhone 12 Pro Max. }
    \label{fig:real_result}
    \vspace{-4mm}
\end{figure}

Fig. \ref{fig:result} and \ref{fig:real_result} shows qualitative results of \OURS on unseen garments for simulated and real data. Following sections discuss the qualitative results and ablation studies.

\mypara{Comparison to Alternative Approaches.}
To the best of our knowledge, there is currently no prior work that
performs our task exactly (i.e., category-level garment pose estimation). To provide baselines for comparison, we consider the following alternative approaches:
(1) \textbf{NN}: retrieve nearest neighbor example in training dataset using global PointNet++ feature, extracted from input point cloud observation.  
(2) \textbf{Direct}: performs shape completion and canonical coordinate labeling in the task space. 

Tab. \ref{tab:baselines} shows that while [NN] can achieve similar Chamfer distance ($D_c$), the pose estimation error ($D_n$) is significantly higher, indicating that the retrieved nearest neighbor mesh does not share a similar configuration as the input observation, while the geometry might be similar. The small gap in Chamfer Distance is because in task space, all the garment surfaces are crumpled and close to each other. Therefore, naively measuring surface distance using closest point correspondence cannot reflect the algorithm's performance on pose estimation. 

Comparison with [Direct] shows the benefit of canonical space representation.  By mapping the partial observation into a canonical space, the network can leverage a stronger shape prior, which is invariant to the garment configurations. Instead, directly performing shape completion in task space requires the network to reason about all possible configurations, which is much more challenging. 

\begin{figure}[t]
    \centering 
    \includegraphics[width=\linewidth]{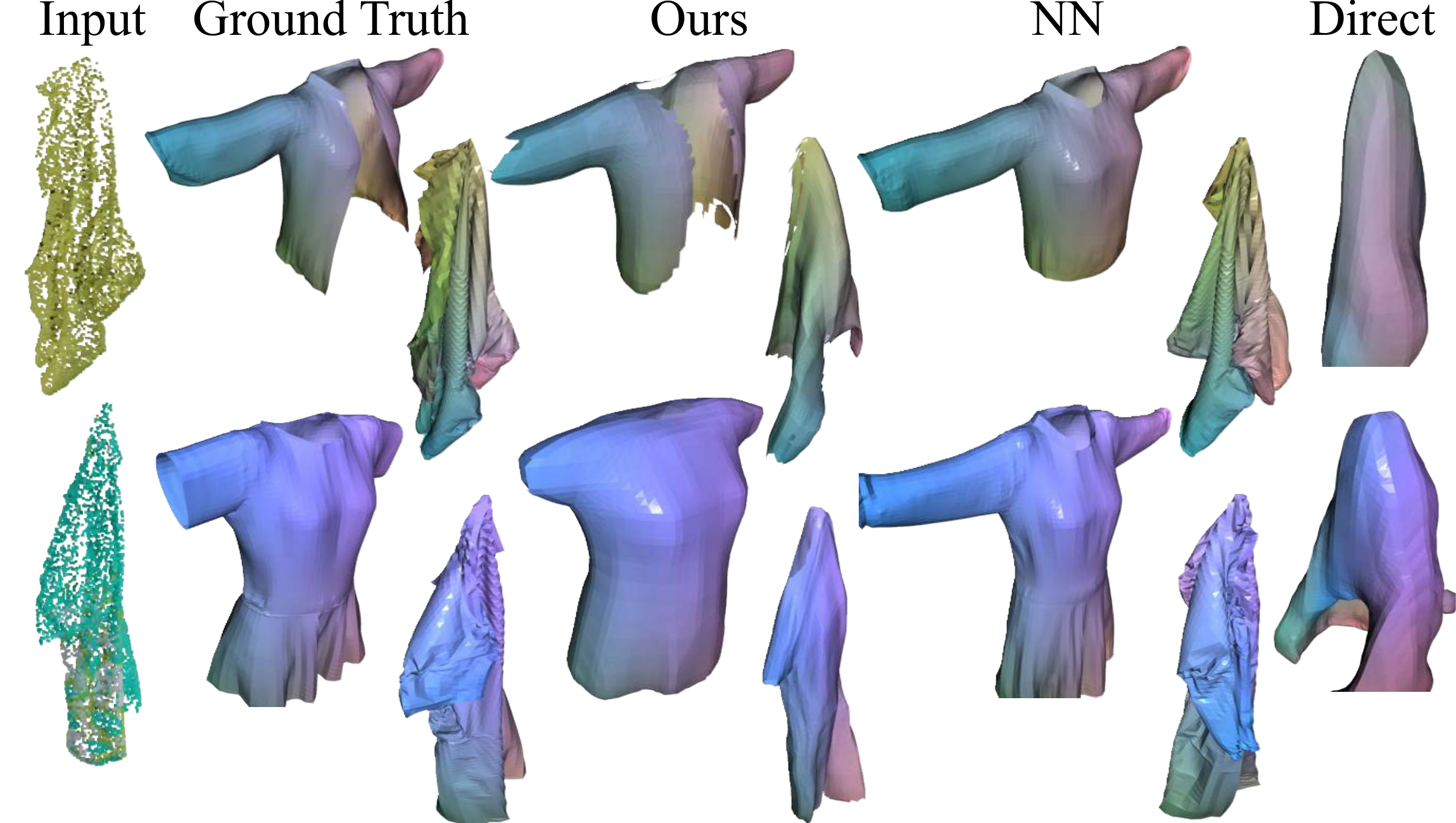}
    \caption{\textbf{Comparison with Alternative Approaches.} NN: nearest neighbor retrieval. Direct: directly perform shape completion and NOCS prediction in the task space.}
    \label{fig:nn_task_space_ours} \vspace{-2mm}
\end{figure} 
\begin{table}
    \centering
     \tabcolsep=0.15cm
     \small 
    \begin{tabular}{l|l|cccccc} 
    \toprule
 & Method & Dress & Jump. & Skirt & Top & Pants & Shirt\\ \midrule 
 $D_c$  & NN  & \textbf{2.09} & 1.89 & 2.18 & 1.82 & \textbf{1.39} & 1.69\\
        & Direct & 13.62 & 63.81 & 12.55 & 9.63 & 11.12 & 9.40 \\
        & Ours & 2.12 & \textbf{1.82} & \textbf{2.14} & \textbf{1.54} & 1.41 & \textbf{1.63}\\\bottomrule
$D_n$  & NN  & 12.74 & 13.38 & 20.55 & 11.57 & 12.43 & 12.11\\
        & Direct  & 48.32 & 81.79 & 39.73 & 31.19 & 36.44 & 43.43 \\
        & Ours & \textbf{6.63} & \textbf{6.06} & \textbf{7.34} & \textbf{4.47} & \textbf{4.37} & \textbf{4.94}\\\midrule 
     
\end{tabular}
\caption{\textbf{Pose Estimation.}  While NN achieves comparable Chamfer distance ($D_c$), the pose estimation error ($D_n$) is significantly higher, indicates that the retrieved mesh does not share similar configuration as the input, while the geometry might be similar.   }\label{tab:baselines} \vspace{-3mm}
\end{table}

\begin{figure}[h]  
       \centering
       \includegraphics[width=\linewidth]{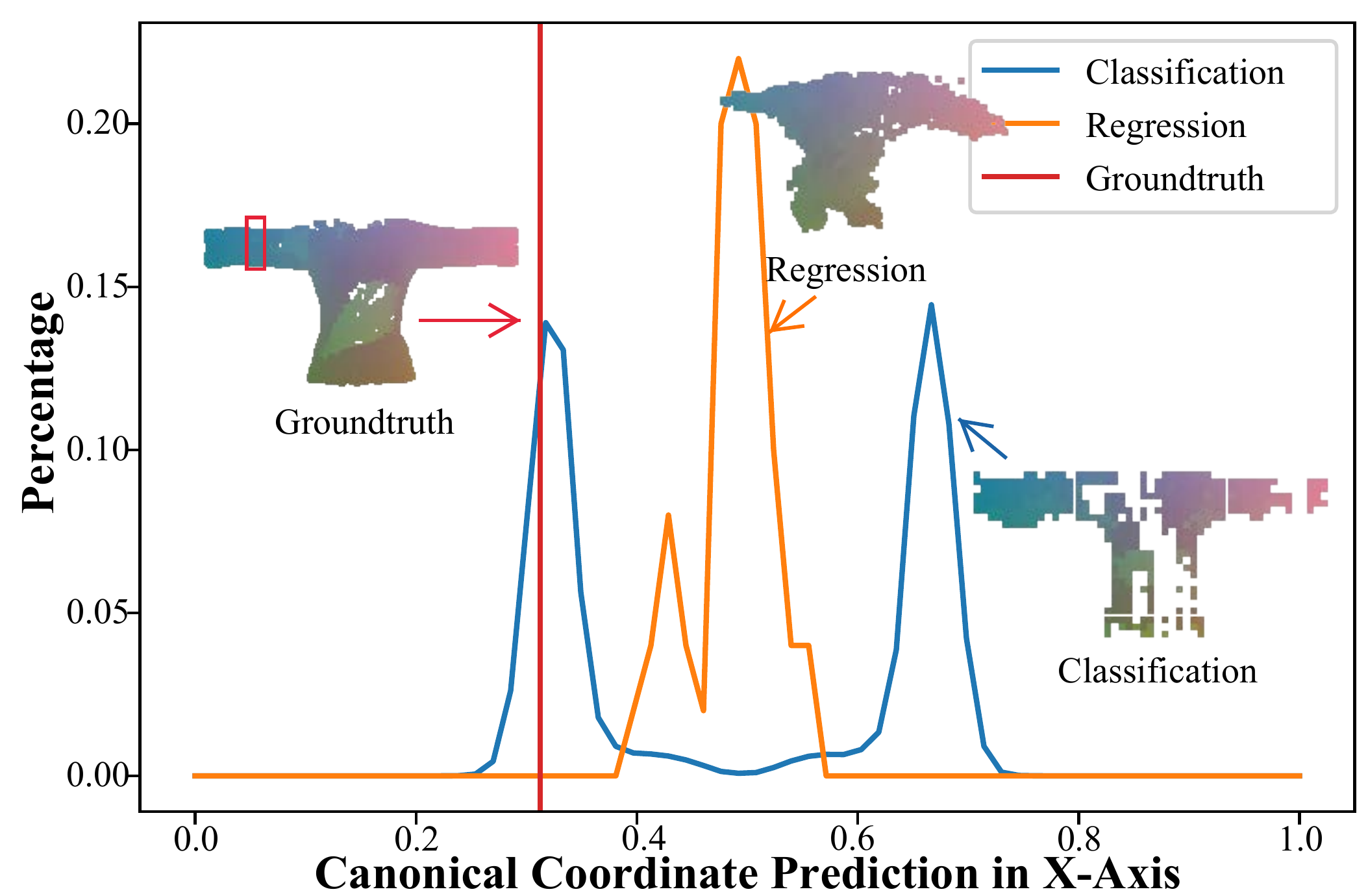} 
       \caption{\textbf{Classification v.s. Regression.} We visualize the canonical coordinate prediction for selected points (i.e., within the red box on the left sleeve). The histogram distribution shows that while classification model predicts a bimodal distribution, the regression model predict a distribution that is close to the mean of the two possible hypotheses, which is far from either solutions.}
       \label{fig:regression_classification}\vspace{-4mm}
\end{figure}

\mypara{NOCS Prediction: Classification v.s. Regression.}
Due to the symmetry in garments, we found that the classification loss is more effective for NOCS prediction than regression loss. As shown in Fig. \ref{fig:regression_classification}, for a set of selected points on the left sleeve, the classification network predicted a bimodal distribution, expressing that the points are likely to be on the same location of either left or right sleeve. In contrast, the regression model predicts coordinates in the middle of two possible locations. This difference is also visible on the final NOCS coordinates visualization, where the regression model tends to map all observations to the middle section of the cloth.
Quantitatively, the classification model also produces lower error in NOCS prediction comparing to the regression model: $0.14$ v.s. $0.16$ (in NOCS space) and $0.06$ v.s. $0.11$ if we consider symmetry in the error computation (i.e., calculating minimal distance between the predicted NOCS and ground truth as well as left-right mirrored ground truth labels). 

\begin{figure}[t]
    \centering
    \includegraphics[width=\linewidth]{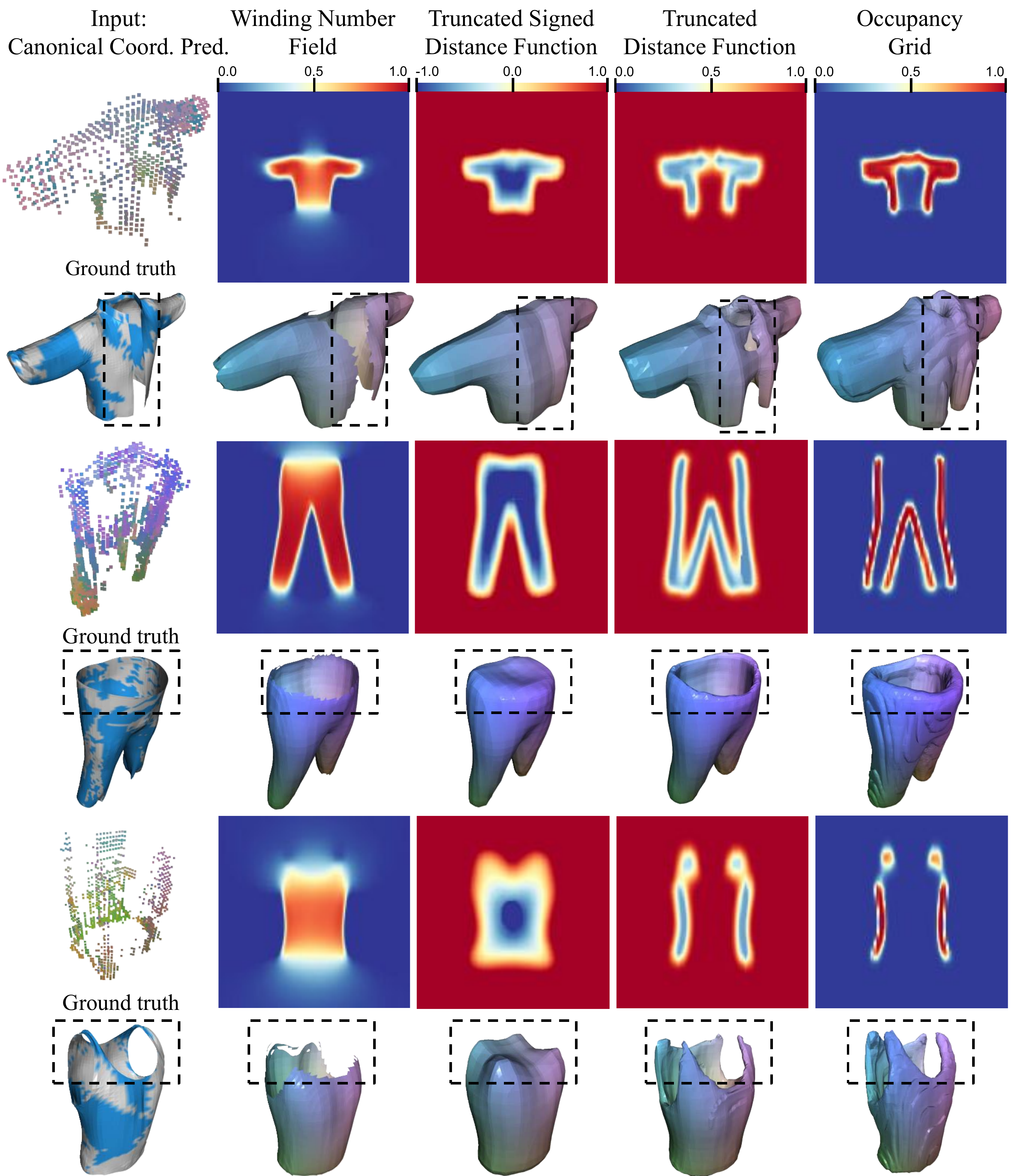}
    \caption{\textbf{Shape Completion.} The magnitude of spatial gradient is used to predict surface openings in winding number field (our representation). TSDF can only represent water-tight surface. TDF predicts thick, over-smoothed surface. Occupancy grid can not represent fabric thinner than the voxel size. On the ground truth mesh, grey indicates visible surface in input, blue indicates occluded surface.}
    \label{fig:completion} \vspace{-3mm}
\end{figure}

\mypara{Shape Representation.}
As shown in Fig. \ref{fig:completion}, using the gradient magnitude of the winding number field, our method is able to predict the front opening of jackets or waist for pants. 
In contrast, TSDF cannot predict the surfaces with openings. The ability to predict opening quantitatively improves our evaluation metrics and informs downstream tasks such as motion planning and physical simulation about the topology of the garment. 
Occupancy grid can represent the cloth with openings as a watertight surface with genus $> 0$. However, its prediction accuracy is limited by volume resolution. While increasing grid resolution improves the surface accuracy, it will result in sparse data distribution that negatively impact the network training. In our experiments, a $128^3$ occupancy grid has an occupancy rate of $0.4\%$, which caused the network to predict only zero value. Therefore, we decrease the occupancy grid resolution to $64^3$ for evaluation. 
Similar to Occupancy grid, Truncated Unsigned Distance Function (i.e. TDF) causes the network to predict over smoothed distance field, result in thick surface prediction and lower performance in most of garment categories. However, the thick surface prediction is better at capturing thin strips on Tops, which the winding number field tends to miss.
Overall, our method yields a $26.7\%$ improvement over TSDF, $18.7\%$ improvement over TDF and $32.2\%$ improvement over occupancy grid.

\begin{table}[h]
    \centering
     \tabcolsep=0.2cm
     \small 
    \begin{tabular}{l|cccccc} 
    \toprule
Method  & Dress & Jumpsuit & Skirt & Top & Pants & Shirt\\ \midrule 
OCC & 2.94 & 3.00 & 2.44 & 1.43 & 2.03 & 2.50\\ 
TSDF & 2.45 & 1.76 & 3.03 & 2.38 & 1.44 & 1.98\\ 
TDF & 2.55 & 2.18 & 2.08 & \textbf{1.22} & 1.67 & 2.11\\
Ours & \textbf{1.94} & \textbf{1.45}  & \textbf{2.00} & 1.30 & \textbf{1.03} & \textbf{1.70} \\ \bottomrule 
\end{tabular}
\vspace{2mm}
\caption{\textbf{Shape Completion Error.} with different shape representations. The error is measured using Chamfer distance (cm) under the canonical pose. Occ: occupancy grid,  TSDF: truncated signed distance function, TDF: truncated unsigned distance function  and Ours: winding number field. }\label{tab:supp_reconstruction} \vspace{-1mm}
\end{table}

\begin{figure}[h]
    \centering
    \includegraphics[width=\linewidth]{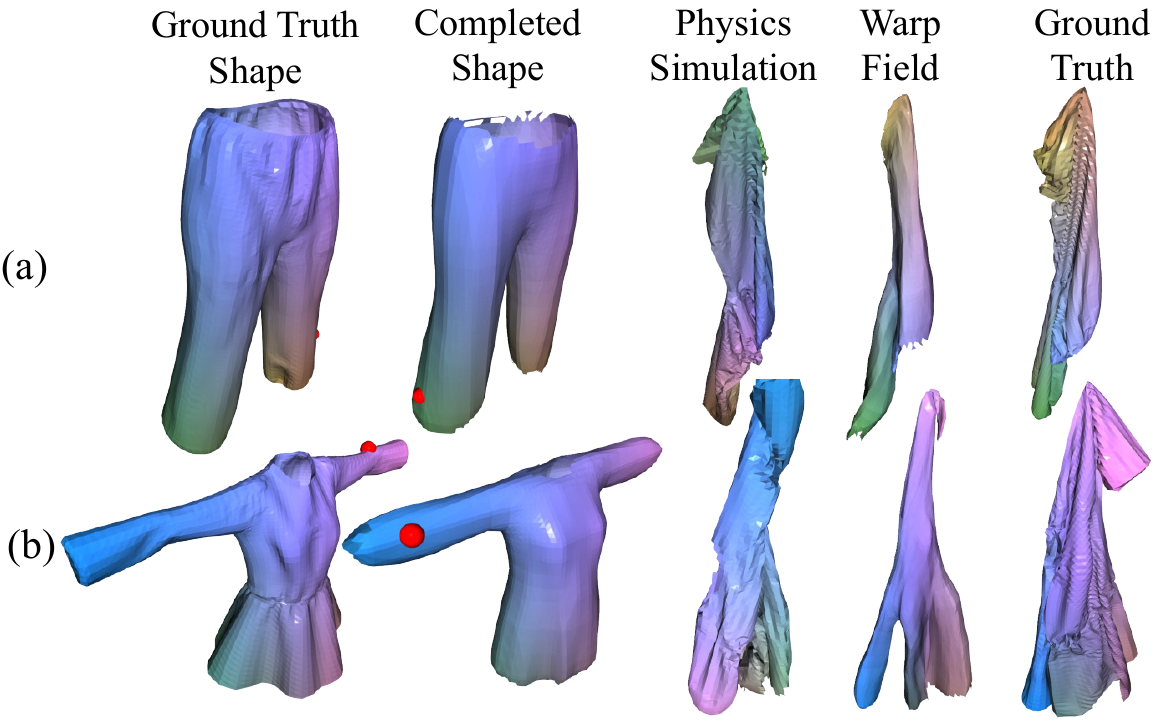}
    \caption{\textbf{Wrap Field Prediction v.s. Physic Simulation.} While physics simulation can always provide physically plausible mapping, wrap field prediction overall generates a more accurate estimation by correcting the errors introduced in earlier steps. For example, the warp field prediction is able to self-correct the mirrored gripping point prediction in both cases (a,b). }
    \label{fig:warp_vs_sim}
\end{figure}

\mypara{Learned Warp Field v.s. Physics Simulation.}
As we discussed in Sec. \ref{sec:warp}, there are two ways to convert the completed 3D mesh back to the task space: (1) using physics simulation with rotation alignment and (2) using a learned implicit warp field prediction network. 
Tab. \ref{tab:warp_vs_sim} and Fig. \ref{fig:warp_vs_sim} shows the comparison of these two approaches.  While physics simulation can generate physically plausible results, the learned warp field often yields more accurate predictions. This is achieved by self-correcting NOCS prediction error in the warp field prediction step using the raw task space point coordinate carried in the feature volume.

\begin{table}[t]
    \centering
     \tabcolsep=0.15cm
     \small 
    \begin{tabular}{l|l|cccccc} 
    \toprule
 & Method & Dress & Jump. & Skirt & Top & Pants & Shirt\\ \midrule 
 $D_c$  & PhysSim  & 2.57 & 2.42 & 2.41 & 3.31 & 1.64 & 1.92\\
        & Ours & \textbf{2.12} & \textbf{1.82} & \textbf{2.14} & \textbf{1.54} & \textbf{1.41} & \textbf{1.63}\\\bottomrule
$D_n$  & PhysSim  & 17.11 & 16.80 & 17.43 & 15.67 & 14.74 & 17.58\\
        & Ours & \textbf{6.63} & \textbf{6.06} & \textbf{7.34} & \textbf{4.47} & \textbf{4.37} & \textbf{4.94}\\ 
\bottomrule 
\end{tabular} 
\vspace{0.5mm}
\caption{\textbf{Canonical to Task Space Transform.} We compared the mapping predicted by the implicit warp field and physic simulation using completed mesh and predicted gripping points. Overall, the implicit warp field is able to predict a more accurate mapping by potentially correct errors produced in earlier steps. } \label{tab:warp_vs_sim} \vspace{-3mm}
\end{table}

\vspace{-4mm}

\paragraph{Testing with Realworld Data.}
In this experiment, we want to validate the algorithm's performance with real world data. To do so, we uses a UR5 robot arm to randomly pick up garment on the table and capture an RGB Point Cloud using an iPhone 12 Pro Max. We filter background points by applying a constant threshold on the x,y,z coordinate.   
Fig. \ref{fig:real_result} shows the qualitative visualization of algorithm prediction.  While the algorithm is trained with simulated data, the model is able to complete 3D geometry for different garment instances and estimate its pose in the grasped state.
Note that all the garments used in this experiment are not presented during training. 

\section{Conclusion}  
We presented \OURS for category-level garments pose estimation from a partial point cloud observation. Experiments demonstrate that GarmentNets is able to generalize to unseen garment instances for both real-world and simulated images and achieve significantly better performance compared to the alternative approaches. 
While achieving promising testing results with real-world data, training \OURS requires a large amount of high-quality data with detailed labels (e.g., pre-aligned 3D meshes and dense correspondence labels), which are all difficult to obtain for the real-world data. 
Future work could consider developing self- or weakly- supervised methods that would allow training such algorithm directly with real-world data without expensive data annotation.

\newpage
\noindent\textbf{Acknowledgement}
The authors would like to thank Eric Cousineau, Benjamin Burchfiel Naveen Kuppuswamy, and other researchers in Toyota Research Institute for their helpful feedback and fruitful discussions.We would also like to thank Huy Ha for data collection, Google for their donation of UR5 robots. This work was supported in part by the Amazon Research Award and the National Science Foundation under CMMI-2037101.

{\small
\bibliographystyle{ieee_fullname}
\bibliography{reference}

\begin{thebibliography}{10}\itemsep=-1pt

\bibitem{fast_wnf}
Gavin Barill, Neil~G. Dickson, Ryan Schmidt, David I.~W. Levin, and Alec
  Jacobson.
\newblock Fast winding numbers for soups and clouds.
\newblock {\em ACM Trans. Graph.}, 37(4), July 2018.

\bibitem{cloth_3d}
Hugo Bertiche, Meysam Madadi, and Sergio Escalera.
\newblock Cloth3d: Clothed 3d humans.
\newblock In Andrea Vedaldi, Horst Bischof, Thomas Brox, and Jan-Michael Frahm,
  editors, {\em Computer Vision -- ECCV 2020}, pages 344--359, Cham, 2020.
  Springer International Publishing.

\bibitem{r1_bhatnagar2020ipnet}
Bharat~Lal Bhatnagar, Cristian Sminchisescu, Christian Theobalt, and Gerard
  Pons-Moll.
\newblock Combining implicit function learning and parametric models for 3d
  human reconstruction.
\newblock In {\em European Conference on Computer Vision ({ECCV})}. {Springer},
  aug 2020.

\bibitem{r1_bhatnagar2020loopreg}
Bharat~Lal Bhatnagar, Cristian Sminchisescu, Christian Theobalt, and Gerard
  Pons-Moll.
\newblock Loopreg: Self-supervised learning of implicit surface
  correspondences, pose and shape for 3d human mesh registration.
\newblock In {\em Neural Information Processing Systems (NeurIPS)}, December
  2020.

\bibitem{cpcpd}
C. {Chi} and D. {Berenson}.
\newblock Occlusion-robust deformable object tracking without physics
  simulation.
\newblock In {\em 2019 IEEE/RSJ International Conference on Intelligent Robots
  and Systems (IROS)}, pages 6443--6450, 2019.

\bibitem{r1_chibane2020ndf}
Julian Chibane, Aymen Mir, and Gerard Pons-Moll.
\newblock Neural unsigned distance fields for implicit function learning.
\newblock In {\em NeurIPS)}, 2020.

\bibitem{unet_3d}
{\"O}zg{\"u}n {\c{C}}i{\c{c}}ek, Ahmed Abdulkadir, Soeren~S. Lienkamp, Thomas
  Brox, and Olaf Ronneberger.
\newblock 3d u-net: Learning dense volumetric segmentation from sparse
  annotation.
\newblock In Sebastien Ourselin, Leo Joskowicz, Mert~R. Sabuncu, Gozde Unal,
  and William Wells, editors, {\em Medical Image Computing and
  Computer-Assisted Intervention -- MICCAI 2016}, pages 424--432, Cham, 2016.
  Springer International Publishing.

\bibitem{shape_completion_encoder}
A. {Dai}, C.~R. {Qi}, and M. {Nießner}.
\newblock Shape completion using 3d-encoder-predictor cnns and shape synthesis.
\newblock In {\em 2017 IEEE Conference on Computer Vision and Pattern
  Recognition (CVPR)}, pages 6545--6554, 2017.

\bibitem{deepgarment}
R. Danundefined\v{r}ek, E. Dibra, C. \"{O}ztireli, R. Ziegler, and M. Gross.
\newblock Deepgarment: 3d garment shape estimation from a single image.
\newblock {\em Comput. Graph. Forum}, 36(2):269–280, May 2017.

\bibitem{imagenet}
J. {Deng}, W. {Dong}, R. {Socher}, L. {Li}, {Kai Li}, and {Li Fei-Fei}.
\newblock Imagenet: A large-scale hierarchical image database.
\newblock In {\em 2009 IEEE Conference on Computer Vision and Pattern
  Recognition}, pages 248--255, 2009.

\bibitem{dense_pose}
R.~A. {Güler}, N. {Neverova}, and I. {Kokkinos}.
\newblock Densepose: Dense human pose estimation in the wild.
\newblock In {\em 2018 IEEE/CVF Conference on Computer Vision and Pattern
  Recognition}, pages 7297--7306, 2018.

\bibitem{resnet}
K. {He}, X. {Zhang}, S. {Ren}, and J. {Sun}.
\newblock Deep residual learning for image recognition.
\newblock In {\em 2016 IEEE Conference on Computer Vision and Pattern
  Recognition (CVPR)}, pages 770--778, 2016.

\bibitem{innmann2016volumedeform}
Matthias Innmann, Michael Zollh{\"o}fer, Matthias Nie{\ss}ner, Christian
  Theobalt, and Marc Stamminger.
\newblock Volumedeform: Real-time volumetric non-rigid reconstruction.
\newblock In {\em European Conference on Computer Vision}, pages 362--379.
  Springer, 2016.

\bibitem{jacobson2013robust}
Alec Jacobson, Ladislav Kavan, and Olga Sorkine-Hornung.
\newblock Robust inside-outside segmentation using generalized winding numbers.
\newblock {\em ACM Transactions on Graphics (TOG)}, 32(4):1--12, 2013.

\bibitem{wnf}
Alec Jacobson, Ladislav Kavan, and Olga Sorkine-Hornung.
\newblock Robust inside-outside segmentation using generalized winding numbers.
\newblock {\em ACM Trans. Graph.}, 32(4), July 2013.

\bibitem{r1_jiang2020bcnet}
Boyi Jiang, Juyong Zhang, Yang Hong, Jinhao Luo, Ligang Liu, and Hujun Bao.
\newblock Bcnet: Learning body and cloth shape from a single image.
\newblock In {\em European Conference on Computer Vision}. Springer, 2020.

\bibitem{Li_2020_CVPR}
Xiaolong Li, He Wang, Li Yi, Leonidas~J. Guibas, A.~Lynn Abbott, and Shuran
  Song.
\newblock Category-level articulated object pose estimation.
\newblock In {\em Proceedings of the IEEE/CVF Conference on Computer Vision and
  Pattern Recognition (CVPR)}, June 2020.

\bibitem{peter_alan_baseline}
Y. {Li}, C. {Chen}, and P.~K. {Allen}.
\newblock Recognition of deformable object category and pose.
\newblock In {\em 2014 IEEE International Conference on Robotics and Automation
  (ICRA)}, pages 5558--5564, 2014.

\bibitem{smpl}
Matthew Loper, Naureen Mahmood, Javier Romero, Gerard Pons-Moll, and Michael~J.
  Black.
\newblock Smpl: A skinned multi-person linear model.
\newblock {\em ACM Trans. Graph.}, 34(6), Oct. 2015.

\bibitem{conv_implicit}
L. {Mescheder}, M. {Oechsle}, M. {Niemeyer}, S. {Nowozin}, and A. {Geiger}.
\newblock Occupancy networks: Learning 3d reconstruction in function space.
\newblock In {\em 2019 IEEE/CVF Conference on Computer Vision and Pattern
  Recognition (CVPR)}, pages 4455--4465, 2019.

\bibitem{dynamic_fusion}
R.~A. {Newcombe}, D. {Fox}, and S.~M. {Seitz}.
\newblock Dynamicfusion: Reconstruction and tracking of non-rigid scenes in
  real-time.
\newblock In {\em 2015 IEEE Conference on Computer Vision and Pattern
  Recognition (CVPR)}, pages 343--352, 2015.

\bibitem{r3_patel20tailornet}
Chaitanya Patel, Zhouyingcheng Liao, and Gerard Pons-Moll.
\newblock Tailornet: Predicting clothing in 3d as a function of human pose,
  shape and garment style.
\newblock In {\em {IEEE} Conference on Computer Vision and Pattern Recognition
  (CVPR)}. {IEEE}, jun 2020.

\bibitem{clothcap}
Gerard Pons-Moll, Sergi Pujades, Sonny Hu, and Michael~J. Black.
\newblock Clothcap: Seamless 4d clothing capture and retargeting.
\newblock {\em ACM Trans. Graph.}, 36(4), July 2017.

\bibitem{pointnet2}
Charles~R. Qi, Li Yi, Hao Su, and Leonidas~J. Guibas.
\newblock Pointnet++: Deep hierarchical feature learning on point sets in a
  metric space.
\newblock In {\em Proceedings of the 31st International Conference on Neural
  Information Processing Systems}, NIPS'17, page 5105–5114, Red Hook, NY,
  USA, 2017. Curran Associates Inc.

\bibitem{r1_saito2020cvpr}
Shunsuke Saito, Jinlong Yang, Qianli Ma, and Michael~J. Black.
\newblock {SCANimate}: Weakly supervised learning of skinned clothed avatar
  networks.
\newblock In {\em Proceedings IEEE/CVF Conf.~on Computer Vision and Pattern
  Recognition (CVPR)}, June 2021.

\bibitem{cpd_physics}
T. {Tang}, Y. {Fan}, H. {Lin}, and M. {Tomizuka}.
\newblock State estimation for deformable objects by point registration and
  dynamic simulation.
\newblock In {\em 2017 IEEE/RSJ International Conference on Intelligent Robots
  and Systems (IROS)}, pages 2427--2433, 2017.

\bibitem{Tekin_2018}
Bugra Tekin, Sudipta~N. Sinha, and Pascal Fua.
\newblock Real-time seamless single shot 6d object pose prediction.
\newblock {\em CVPR}, 2018.

\bibitem{wang2019densefusion}
Chen Wang, Danfei Xu, Yuke Zhu, Roberto Martín-Martín, Cewu Lu, Li Fei-Fei,
  and Silvio Savarese.
\newblock Densefusion: 6d object pose estimation by iterative dense fusion.
\newblock {\em CVPR}, 2019.

\bibitem{Wang_2019}
He Wang, Srinath Sridhar, Jingwei Huang, Julien Valentin, Shuran Song, and
  Leonidas~J. Guibas.
\newblock Normalized object coordinate space for category-level 6d object pose
  and size estimation.
\newblock {\em CVPR}, 2019.

\bibitem{nocs}
H. {Wang}, S. {Sridhar}, J. {Huang}, J. {Valentin}, S. {Song}, and L.~J.
  {Guibas}.
\newblock Normalized object coordinate space for category-level 6d object pose
  and size estimation.
\newblock In {\em 2019 IEEE/CVF Conference on Computer Vision and Pattern
  Recognition (CVPR)}, pages 2637--2646, 2019.

\bibitem{whelan2016elasticfusion}
Thomas Whelan, Renato~F Salas-Moreno, Ben Glocker, Andrew~J Davison, and Stefan
  Leutenegger.
\newblock Elasticfusion: Real-time dense slam and light source estimation.
\newblock {\em The International Journal of Robotics Research},
  35(14):1697--1716, 2016.

\bibitem{marked_cloth}
Ryan White, Keenan Crane, and D.~A. Forsyth.
\newblock Capturing and animating occluded cloth.
\newblock In {\em ACM SIGGRAPH 2007 Papers}, SIGGRAPH '07, page 34–es, New
  York, NY, USA, 2007. Association for Computing Machinery.

\bibitem{Xiang_2018}
Yu Xiang, Tanner Schmidt, Venkatraman Narayanan, and Dieter Fox.
\newblock Posecnn: A convolutional neural network for 6d object pose estimation
  in cluttered scenes.
\newblock {\em RSS}, 2018.

\bibitem{xiao2005uncalibrated}
Jing Xiao and Takeo Kanade.
\newblock Uncalibrated perspective reconstruction of deformable structures.
\newblock In {\em Tenth IEEE International Conference on Computer Vision
  (ICCV'05) Volume 1}, volume~2, pages 1075--1082. IEEE, 2005.

\bibitem{pcn}
W. {Yuan}, T. {Khot}, D. {Held}, C. {Mertz}, and M. {Hebert}.
\newblock Pcn: Point completion network.
\newblock In {\em 2018 International Conference on 3D Vision (3DV)}, pages
  728--737, 2018.

\bibitem{Zakharov_2019}
Sergey Zakharov, Ivan Shugurov, and Slobodan Ilic.
\newblock Dpod: 6d pose object detector and refiner.
\newblock {\em ICCV}, 2019.

\bibitem{zeng2018robotic}
Andy Zeng, Shuran Song, Kuan-Ting Yu, Elliott Donlon, Francois~R Hogan, Maria
  Bauza, Daolin Ma, Orion Taylor, Melody Liu, Eudald Romo, et~al.
\newblock Robotic pick-and-place of novel objects in clutter with
  multi-affordance grasping and cross-domain image matching.
\newblock In {\em 2018 IEEE international conference on robotics and automation
  (ICRA)}, pages 3750--3757. IEEE, 2018.

\bibitem{deep_fasion3d}
Heming Zhu, Yu Cao, Hang Jin, Weikai Chen, Dong Du, Zhangye Wang, Shuguang Cui,
  and Xiaoguang Han.
\newblock Deep fashion3d: A dataset and benchmark for 3d garment reconstruction
  from single images.
\newblock In Andrea Vedaldi, Horst Bischof, Thomas Brox, and Jan-Michael Frahm,
  editors, {\em Computer Vision -- ECCV 2020}, pages 512--530, Cham, 2020.
  Springer International Publishing.

\end{thebibliography}
}

\clearpage
\appendix

\section{Network Architecture Details}
\paragraph{Canonical Coordinate Prediction Network}
To predict per-point canonical coordinate, we used a PointNet++ \cite{pointnet2} network in Multi-Resolution Grouping (MRG) configuration. The network consists of 3 Set Abstraction layers and 3 Feature Propagation layers. Detailed parameters are shown in Tab. \ref{tab:sa_params}. The final per-point 128 dimensional feature vector is transformed with a 3-layer MLP to perform $3\times 64$ way classification with Cross Entropy Loss.

\begin{table}[h]
\centering
     \tabcolsep=0.08cm
     \small 
\begin{tabular}{l|ccccc}
\toprule
Layer & SA Radius & SA Ratio & SA Features & FP k & FP Features \\ \hline
1     & 0.05      & 0.5             & 128         & 3    & 128         \\
2     & 0.1       & 0.25            & 256         & 3    & 128         \\
3     & Inf       & 1               & 1024        & 1    & 256        \\
\bottomrule
\end{tabular}
\vspace{+2mm}
\caption{
    \textbf{PointNet++ Parameters.} SA: parameters for Set Abstraction layers. FP: parameters for Feature Propagation layers.
}
\label{tab:sa_params}
\vspace{-3mm}
\end{table}

\mypara{Feature Completion Network (3D CNN)}
To transform the sparse feature volume scattered from per-point features to a dense feature volume, we used a symmetrical 3D UNet \cite{unet_3d} architecture with 4 levels of encoder/decoder pairs. Each level of encoder/decoder has 32 feature maps.

\mypara{Shape Completion Network}
To predict Winding Number Field (WNF), the interpolated features from dense feature volume is transformed using a 3 layer MLP with feature dimensions [512, 512, 1].

\mypara{Warp Field Network}
Similar to the Shape Completion Network, the interpolated features are transformed using a 3 layer MLP  with feature dimensions [512 ,512, 3].

\section{Additional Results}
\begin{figure*}[t]
    \centering
    \includegraphics[width=0.85\linewidth]{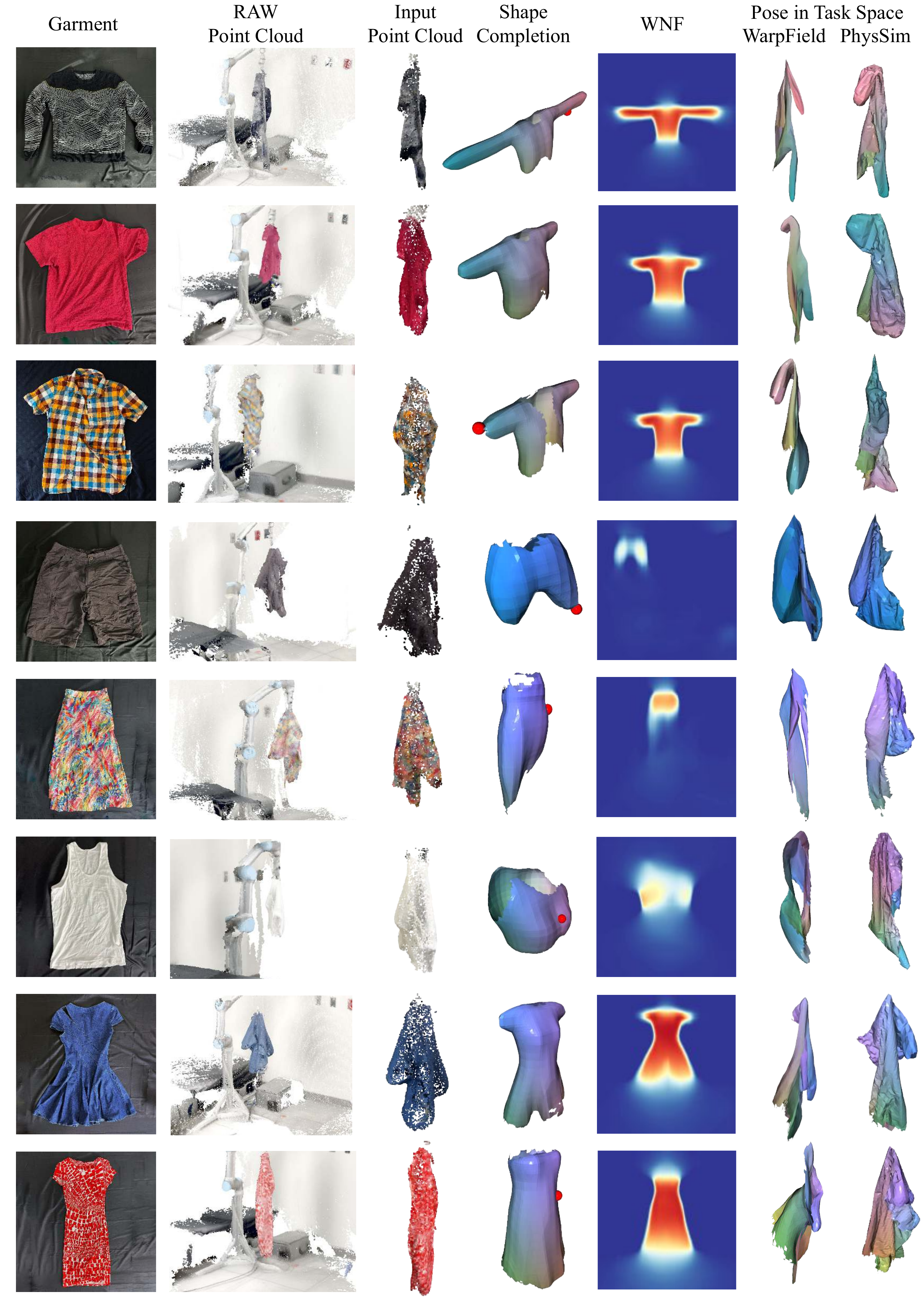}
    \caption{\textbf{Qualitative Results on Unseen Garment Instances (Real World).} }
    \label{fig:supp_real_result}
\end{figure*}

\begin{figure*}[t]
    \centering \vspace{-5mm}
    \includegraphics[width=\linewidth]{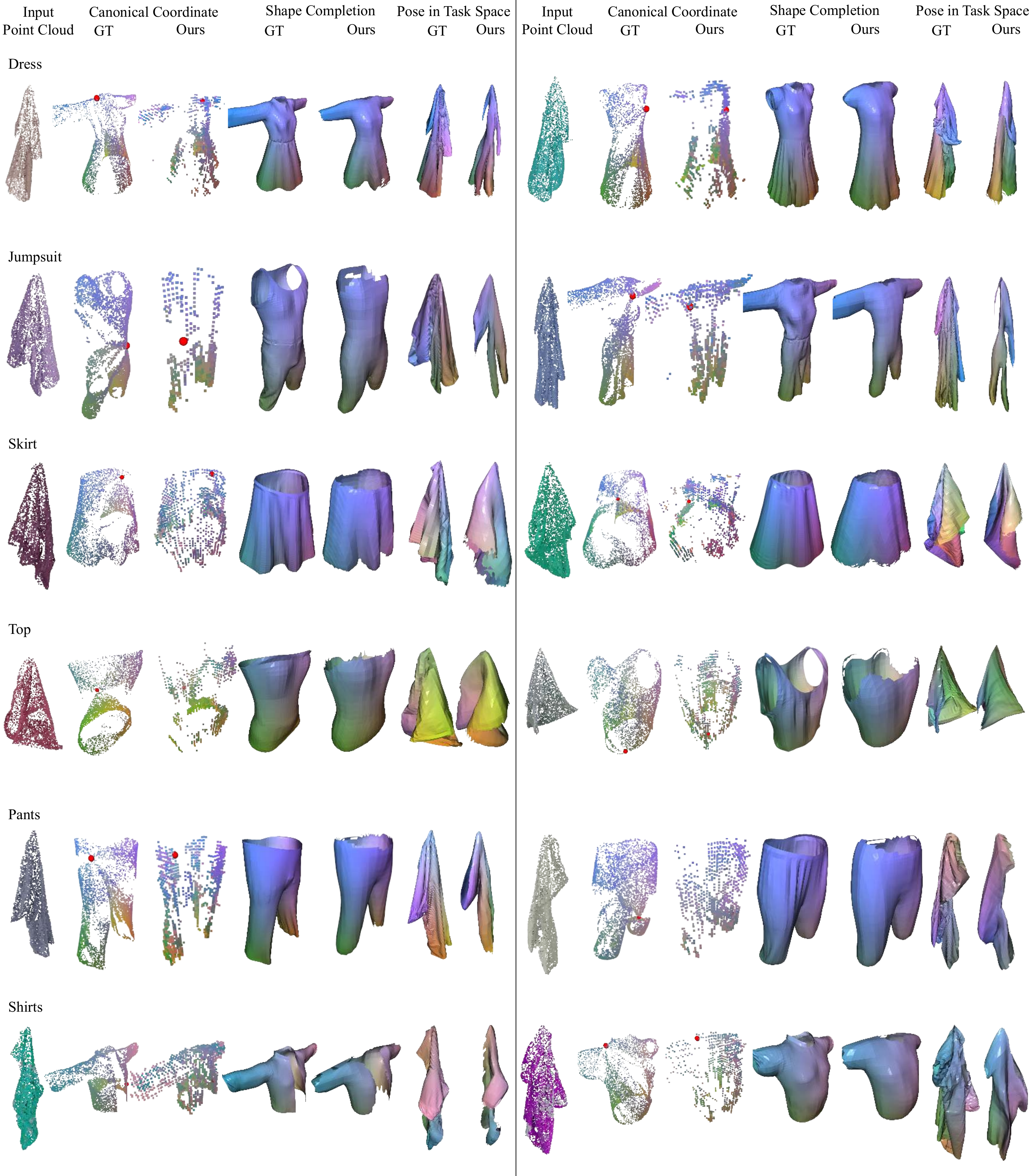}
    \caption{\textbf{Qualitative Results on Unseen Garment Instances (Simulation).} }
    \label{fig:supp_result_all}
    \vspace{-5mm}
\end{figure*}

Fig. \ref{fig:supp_real_result} and \ref{fig:supp_result_all}  show additional  results on real world and simulated data respectively. The real world point cloud are collecting using an iPhone 12 Pro Max. 

\mypara{Garment Category Classification}
Our algorithm described in the paper assumes known garment category for the input point cloud. When dealing with a mixed pile of garments, we assume that the category can be inferred using a classifier.

To validate this assumption, we trained a simple image classifier using only RGB images. The model uses an ImageNet \cite{imagenet} pre-trained ResNet-50 \cite{resnet} backbone to extract a 2048 dimensional feature. The feature is then transformed using a 3-layer MLP to perform 6 way classification with Cross Entropy Loss.

The classifier is trained on each view independently.
During prediction, we use the majority ensemble of all 4 views. This simple model yields $93.85\%$ prediction accuracy on the test set. The confusion matrix is shown in Tab. \ref{tab:confusion_matrix}.

\begin{table}[h]
    \centering
     \tabcolsep=0.15cm
     \small 
\begin{tabular}{l|cccccc}
\toprule
         & Dress & Jumpsuit & Skirt & Top   & Trousers & Tshirt \\ \hline
Dress    & 0.966 & 0.020    & 0.003 & 0.001 & 0.005    & 0.005  \\
Jumpsuit & 0.003 & 0.956    & 0.000 & 0.000 & 0.010    & 0.008  \\
Skirt    & 0.162 & 0.010    & 0.778 & 0.013 & 0.030    & 0.006  \\
Top      & 0.001 & 0.004    & 0.002 & 0.979 & 0.009    & 0.004  \\
Trousers & 0.010 & 0.025    & 0.005 & 0.009 & 0.944    & 0.008  \\
Tshirt   & 0.021 & 0.026    & 0.003 & 0.026 & 0.058    & 0.866 \\
\bottomrule 
\end{tabular}
\vspace{+2mm}
\caption{
    \textbf{Confusion Matrix for Image Classification.}
}
\label{tab:confusion_matrix}
\vspace{-4mm}
\end{table}

\begin{figure*}[t]
    \centering \vspace{-5mm}
    \includegraphics[width=\linewidth]{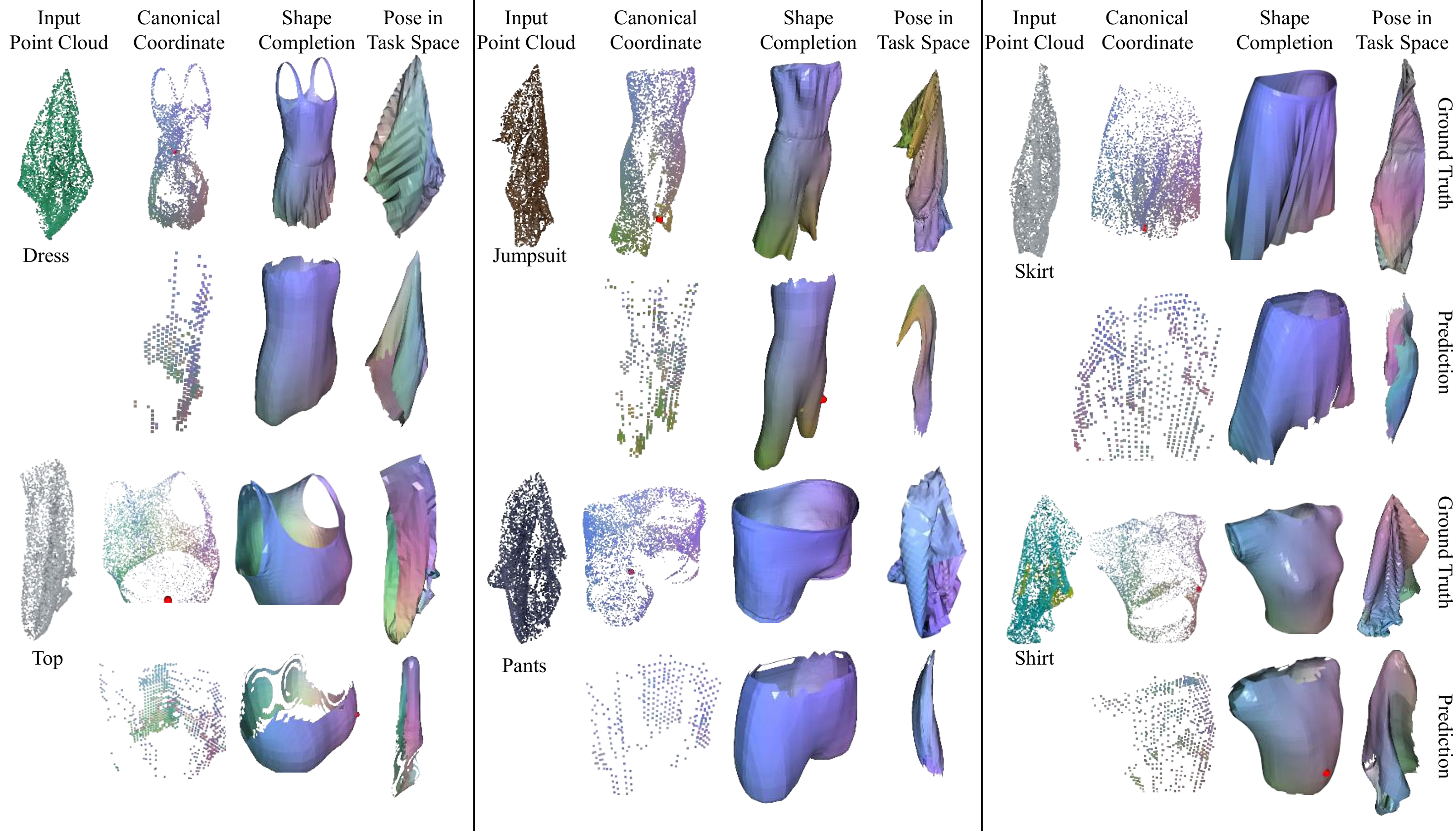}
    \caption{\textbf{Failure cases on Unseen Garment Instances (Simulation).} }
    \label{fig:supp_sim_fail}
    \vspace{-5mm}
\end{figure*}

\mypara{Failure mode analysis}
Fig. \ref{fig:supp_sim_fail} shows various failure cases on unseen simulation data. Due to low sharpness in the predicted winding number field, the canonical reconstruction for the Top and Shirt example have missing faces around the shoulder area. The Jumpsuit, Skirt and Pants example have over-smoothed warp field prediction, resulting in inaccurate task space mesh. The Dress example has missing shoulder strap due to winding number field's inability to represent wire-like structure.

\begin{figure}[t]
\vspace{-2mm}
\begin{center}
    \includegraphics[width=0.9\linewidth]{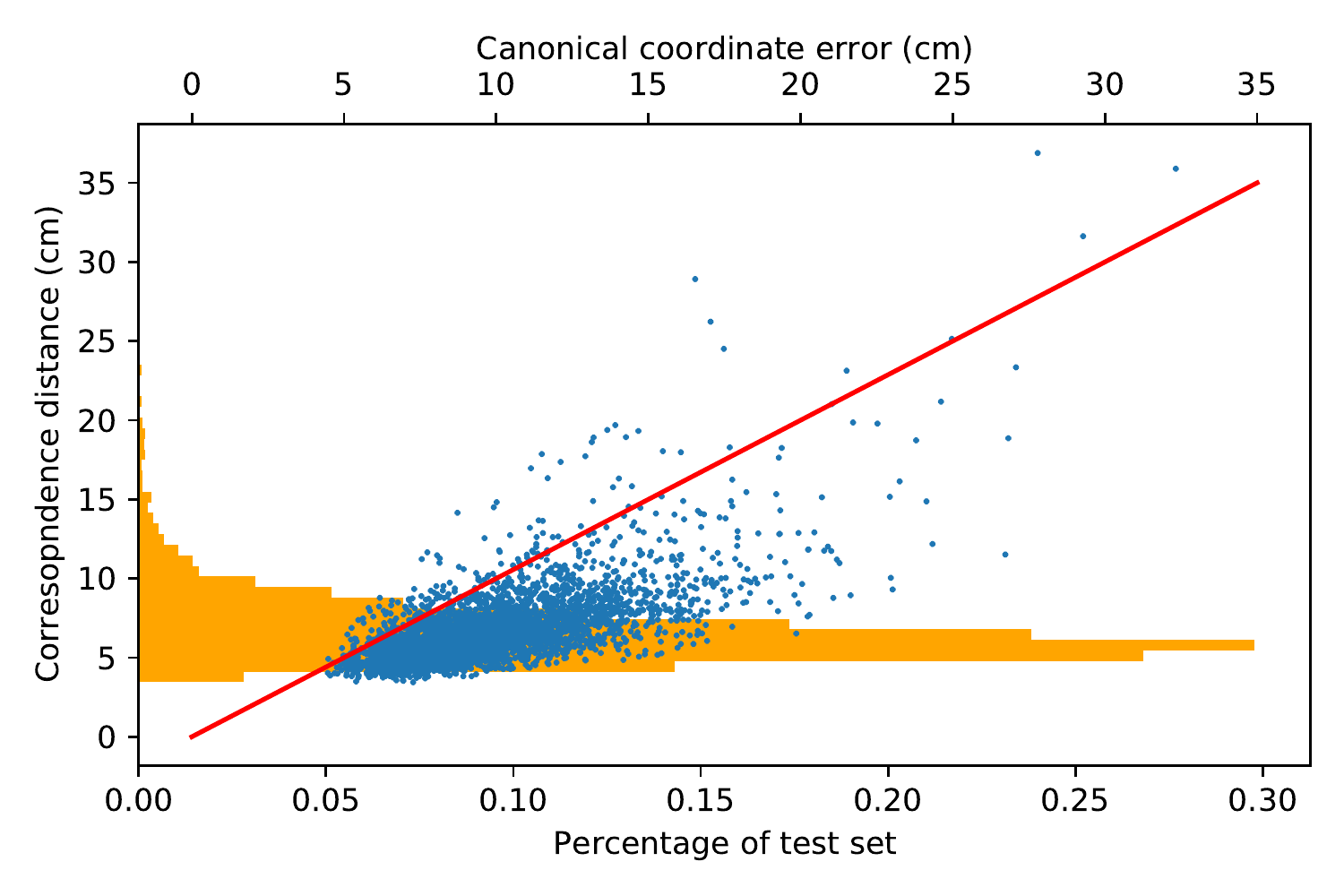}
\end{center} \vspace{-5mm}
   \caption{\textbf{Error distribution and correlation.} The correspondence distance vs canonical coordinate prediction error for Dress category is shown in blue dots. The line of slope 1 is shown in red. The correspondence distance error histogram is shown in orange.} 
\label{fig:nocs_vs_warp}
\end{figure}

\mypara{Error distribution and correlation}
As shown in Fig. \ref{fig:nocs_vs_warp}, the correspondence error is highly correlated to the canonical coordinate error. This suggests that jointly optimizing for both metrics might yield performance improvement.

\mypara{Training Testing Split}
We use CLOTH3D dataset \cite{cloth_3d} for data generation. Tab. \ref{tab:supp_dataset_split} shows the number of garment instances in training testing split for each category. 

\begin{table}[h]
    \centering
     \tabcolsep=0.15cm
     \small 
    \begin{tabular}{l|cccccc}
    \toprule
                & Dress & Jumpsuit & Skirt & Top & Trousers & Tshirt \\ \midrule
    training & 1631  & 1825     & 376   & 840 & 1353     & 889    \\
    validation   & 203   & 227      & 46    & 104 & 169      & 111    \\
    testing  & 203   & 227      & 46    & 104 & 169      & 111    \\ 
    \bottomrule 
    \end{tabular}
    \vspace{2mm}
    \caption{ \textbf{Training, Validation and Testing Split. }
        The number of garment instances used for each category. Each garment instance is simulated 21 times using randomly selected gripping point.
    }
    \label{tab:supp_dataset_split}
\end{table}

\section{Limitations and Future Work}
\cheng{\OURS demonstrates promising result on real-world data while being trained only on synthetic data. However, the inability to propagate gradients from shape completion and warp field prediction modules to the canonical coordinate prediction module prevents us form training end-to-end. More specifically, the correspondence error is highly correlated to the canonical coordinate error, which suggests that jointly optimizing for both metrics might yield performance improvement. This limitation also requires us to manually define a dense correspondence from input to the canonical space, which is expensive to obtain on real-world data.}

\end{document}